# Unsupervised Video Depth Estimation Based on Ego-motion and Disparity Consensus

Lingtao Zhou, Jiaojiao Fang, Guizhong Liu
School of Electronic and Information Engineering Xi'an Jiaotong University, China
zhoulingtao7458@hotmail.com, 995541569@qq.com, liugz@xjtu.edu.cn

*Abstract*—Unsupervised learning based depth estimation methods have received more and more attention as they do not need vast quantities of densely labeled data for training which are touch to acquire. In this paper, we propose a novel unsupervised monocular video depth estimation method in natural scenes by taking advantage of the state-of-the-art method of Zhou et al. which jointly estimates depth and camera motion. Our method advances beyond the baseline method by three aspects: 1) we add an additional signal as supervision to the baseline method by incorporating left-right binocular images reconstruction loss based on the estimated disparities, thus the left frame can be reconstructed by the temporal frames and right frames of stereo vision; 2) the network is trained by jointly using two kinds of view syntheses loss and left-right disparity consistency regularization to estimate depth and pose simultaneously; 3) we use the edge aware smooth L2 regularization to smooth the depth map while preserving the contour of the target. Extensive experiments on the KITTI autonomous driving dataset and Make3D dataset indicate the superiority of our algorithm in training efficiency. We can achieve competitive results with the baseline by only 3/5 times training data. The experimental results also show that our method even outperforms the classical supervised methods that using either ground truth depth or given pose for training.

## I. INTRODUCTION

DEPTH estimation aims to obtain a representation of the 3D spatial structure of a scene, in other words, to obtain the distance of each point of the scene from the image plane. This is one of the fundamental problems in computer vision perception with numerous applications such as autonomous driving, visual navigation, grasping in robotics, robot assisted surgery, and automatic 2D to 3D conversion in film [18], and has a long history of research. Although the traditional methods using geometric learning have been studied for many years, they still failed to model the ability of human to infer the 3D structure of a scene. With the rise of deep learning, there has been a surge in the number of works to learn the regularities of the scene from large amounts of data by the supervised manner. Once the model is trained, it has the scene perception capabilities even from a single image. But these methods are mainly based on the assumption that a plentiful of densely labeled depth data are available which is usually impractical in many applications. Hence, some researchers have attempted to overcome this problem by unsupervised learning, but the existing methods still have some limitations such as the predicted depth maps need global error compensation and are often over smoothing.

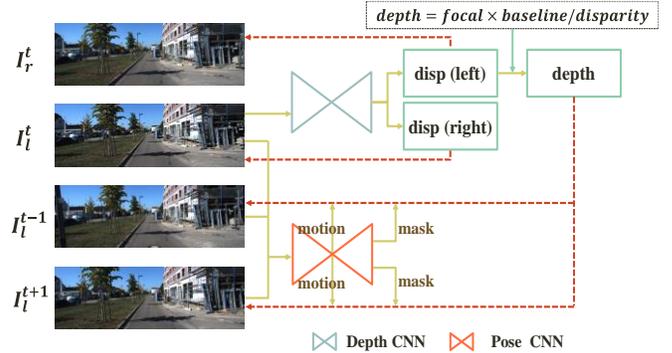

**Figure 1.** Overview of the proposed pipeline for depth estimation and camera pose estimation. The depth network takes only the target view as input with the right binocular image for reconstruction supervision, and outputs a dense per-pixel depth map. The pose network takes both the target view and the nearby source views as input, and outputs the relative poses of camera from target image to the source images.

To learn a mapping function from a monocular image to depth map by unsupervised manner is challenging due to the high uncertainty and dense continuous-valued outputs, so more constraints are needed to solve this problem. In this paper, we propose a novel method for video depth estimation by adding another view synthesis to the existing methods and enhancing the network architecture. Our method relies on the geometrical relationship both about the disparities of stereo vision and structure from ego-motion. We jointly train a monocular depth estimation CNN and camera pose estimation CNN by unlabeled video sequences and the rectified opposite view of stereo vision. For depth estimation, different from the existing methods, we use the warping loss as supervision on temporal frames reconstruction, the disparity based left-right stereo images reconstruction and the consistency constraint of the left-right disparities estimation cooperatively to train the networks. By warping one of the stereo images to match the opposite, we can avoid defining a scale factor for global error compensation of the predicted depth as in existing methods and reduce the influence of the inaccuracy pose on the depth estimation. By adding an edge aware smoothing regularization term, the estimated depth has better contour for foreground object and a significant accuracy improvement of the global depth estimation. As the calibrated right images of stereo vision is also used as supervision to train our model, thus only 3/5 times of samples are available for training. Our trained model can be applied in the case that only monocular images are acquired and achieve competitive results with the baseline algorithm by fewer training data. Experimental results indicate that the proposed method significantly increases the quality of the predicted depth map. Fig. 1 shows the overview of our proposed pipeline for unsupervised depth and camera motion estimation.

Specifically, our main contributions are as follows:

1) we incorporate the disparities based left-right images reconstruction loss as supervision to the existing ego-motion based unsupervised video depth estimation algorithm. By cooperating more additional constraints to the densely per-pixel outputs task, we can avoid the need of global depth error compensation and reduce the number of training samples that need to converge to the global optimal solution;

2) an evaluation on a challenging autonomous driving dataset shows that our method can achieve better results than the baseline method under the condition that only half of training samples are used for training;

3) in addition, we also test our model trained by a challenging autonomous driving dataset on a new outdoor urban dataset to show the generalization ability of the trained model.

## II. RELATED WORK

There are a large amount of studies that focus on depth estimation from images or videos based on deep neural networks, and have made great progress. These methods mainly fall into two categories, supervised methods and unsupervised methods. Although the supervised methods have achieved satisfactory performance, the annotated ground-truth data are laborious to prepare, especially for the dense output task, and usually not available in many real scenarios. While the existing unsupervised methods that based on the multi-views images or predictable camera pose are typically more applicable. Here we mainly focus on the unsupervised methods and some supervised methods that related to monocular depth estimation from videos.

### A. Supervised Single View Image Depth Estimation

Single-view depth estimation refers to the problem setup where only a single image is available for validation.

Saxena et al. [1] propose the famous Make3D model that using the patches over-segmented from the input image to estimate the 3D location and orientation of local planes for explaining each patch. Liu et al. [2] use a convolutional neural network (CNN) to learn the unary and pairwise terms instead of hand-tuning them. Karsch et al. [3] attempt to produce more coherent image level predictions by copying whole depth images from a training set. A drawback of this approach is that it requires the entire training set to be available at test time. Eigen et al. [4] introduce a two scales supervised deep network trained by ground-truth depth to produce dense pixel depth estimates and to learn a representation directly from the raw pixel values. Several works have built upon the success of this approach to further improve accuracy such as changing the loss from regression to classification [5] or using other more robust loss functions [6]. But the superior performance of these approaches mostly rely on having high quality and pixel aligned ground truth depth for training.

### B. Unsupervised Depth Estimation

Recently, a handful of deep network based unsupervised depth estimation methods have been proposed, which do not need ground truth depth during training. This is highly under-constrained and thus these methods perform poorly. The unsupervised depth estimation methods are mainly based on the new view synthesis. Flynn et al. [7] introduce a novel image synthesis network called Deep Stereo that generates new views by selecting pixels from nearby images. During training, the relative pose of multiple cameras is used to predict the appearance of a nearby image. Then the most appropriate depths are selected to sample colors from the neighboring images, based on plane sweep volumes. At test time, image synthesis is performed on small overlapping patches. As it requires several nearby posed images at test time Deep Stereo is not suitable for monocular depth estimation. Xie et al. [8] also propose the Deep3D network to address the problem of novel view synthesis, where their goal is to generate the opposite right view of the binocular pairs from an input left image and is trained by using an image reconstruction loss and stereoscopic film footage as training data. Their method produces a distribution over all the possible disparities for each pixel. The disadvantage of their image formation model is that it increases the number of candidate disparity values greatly increases the memory consumption of the algorithm, making it difficult to scale their approach to bigger output resolution. Like Deep3D, Garg et al. [9] train a network for monocular depth estimation using an image reconstruction loss. However, their objective function is very challenging to optimize. A similar approach is taken by Godard et al. [10], with the addition of a left-right consistency constraint, using bilinear sampling to generate images, resulting in a fully differentiable training loss and a better architecture design that led to impressive performance. In this work, we perform a comparison to the Deep3D image formation model, and show that our algorithm produces superior results.

### C. Unsupervised depth prediction from video

Another line of related works to ours are visual representation learning from video, where the general goal is to design reasonable tasks for learning generic visual features from video data, such as ego-motion estimation[11,12], and inter-frame image reconstruction [13].

Zhou et al. [14] propose a method jointly learning of ego-motion and depth from unlabeled videos by unsupervised manner with the static scene assumption. They learn an interpretable mask to account for local moving. However, these masks are not useful for motion segmentation as they always failed for occlusion at boundaries. Yin et al. [15] add a refinement network to [14] for also estimating residual optical flow and use forward-backward consistency to account for moving regions, but there is no coupling of the optical flow network with the depth and camera pose estimation networks. Mahjourian et al. [16] use a more explicit geometric loss to jointly learn depth and camera motion for rigid scenes. Ranjan et al. [17] introduce a framework that coupled motion segmentation, flow, and depth and camera motion models together and solved jointly to reason the complete geometric structure and motion of the scene. Moniz et al. [18] present an unsupervised approach to estimate 3D facial structures from a single image and 3D viewpoint transformations that match a desired pose and facial geometry. The estimated depth are used as intermediate computations within a new back-propagative loss to predict the parameters of a 3D affine transformation matrix.

In this paper, we also treat monocular depth estimation and camera motion estimation as a new view synthesis problem that simultaneously inferring the scene geometry and the camera ego-motion, thus the disparity can be solved without requiring ground truth label. We synthesis the target image both by temporally frame and its opposite image of stereo vision. And the learned feature representation by our deep neural network (especially the single-view depth CNN) should capture some levels of semantics that could generalize to other scenes. However, only minimizing photometric loss can result in high image reconstructions quality but low quality depth, thus we add additional terms to fully differentiable training loss including a left-right consistency loss, mask regularization as in [14] and edge aware L2 smoothness loss to improve the performance of depth estimation.

III. METHOD

In this section, we will describe the disparities consensus network, pose estimation network, the co-training loss functions and how do we incorporate the disparity consensus into the pose estimation without supervision. In this way, we can get a higher accuracy depth without additional global error compensation. During training we use the opposite image of stereo vision to further constraint the depth estimation, but once the model is trained it can be used to estimate the depth of the dynamic scene only by a single monocular image. Once the network is trained, it can be used separately for depth prediction and pose estimation.

A. *Novel view synthesis based on both spatial and temporal Geometry Optics*

The key supervision signal for our CNNs model comes from the task of novel view synthesis: given one input view of a scene, synthesize an image of the scene seen from a different camera pose or different camera.

Given a training video sequence collected by the left camera,

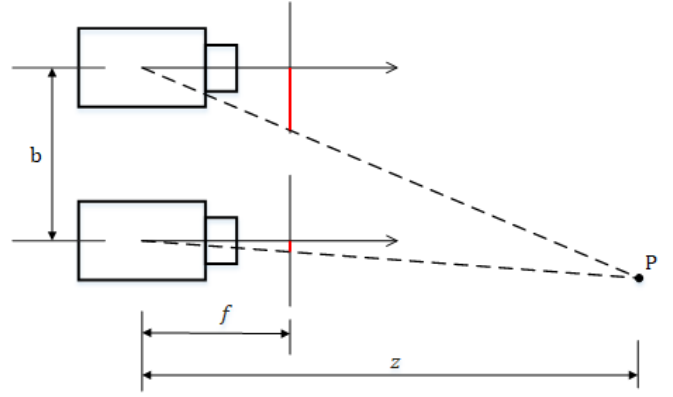

**Figure 2.** Illustration of the explanations for the disparity d which is caused by the two different imaging planes projected by the same space point P. The two red lines indicate the distance from the projected points to the camera center.

which are denoted as $\langle I_l^1, I_l^2, \cdots, I_l^t, \cdots, I_l^N \rangle$ and its rectified stereo counterpart view images $\langle I_r^1, I_r^2, \cdots, I_r^t, \cdots, I_r^N \rangle$ that are all captured by the same camera at the same time, and N is the length of the sequence. We choose the $I_l^t$ as the target image and the rest as the source images which are indicated as $I^s(p)$, then we can use the source images to reconstruct the target image by the following two transformations, the first one is the spatial transformation by the stereo disparity that can reconstructed one image from the other, then the depth information can be learned by:

$$d/b = f/z \qquad (1)$$

Where $d$ corresponds to the image disparity, $b$ is the baseline distances between the two cameras, $f$ is the camera focal length, and $z$ is the depth of the scene. It is intuitive that for the calibrated camera, once the disparity $d_l^t$ of $I_l^t$ is estimated, we can trivially get the 3D depth information of the scene, so here we try to find the disparity caused by two calibrated images instead of predicting the depth of a scene directly. This goal can be easily fulfilled by reconstructing image $\hat{I}_l^t$ by the calibrated image $I_r^t$, and vice versa. That is, we can simultaneously predict left and right disparities by only using the left image $I_l^t$ as input. And we can get the disparity based stereo pair images reconstruction loss as supervision by

$$\mathcal{L}_{ap}^t = |I_l^t - \hat{I}_l^t| + |I_r^t - \hat{I}_r^t|. \qquad (2)$$

Where $\hat{I}_l^t$ is synthesized by the source image $I_r^t$ and left disparity $d_l^t$, $\hat{I}_r^t$ by the target image $I_l^t$ and the right disparity $d_r^t$. This means that given one input image to the network, it can output two different disparity maps by rebuilding different goals.

Due to the dense per pixel disparity prediction being a hard problem, so one more constraint is need to get a better result:

$$\mathcal{L}_{lr}^t = |d_l^t - \hat{d}_l^t| + |d_r^t - \hat{d}_r^t| \qquad (3)$$

where $d_l^t$ and $d_r^t$ are predicted by the neural network using reconstruction loss as supervision, $\hat{d}_l^t$ and $\hat{d}_r^t$ are computed by the corresponding disparity map. We also estimate the right disparity $d_r^t$ during training which would not be used at test time. The disparity is predicted by warping one image to another. This is reasonable as the disparity is easy to achieve from a pair of rectified stereo images.

Another is the temporal geometric warping that is based on

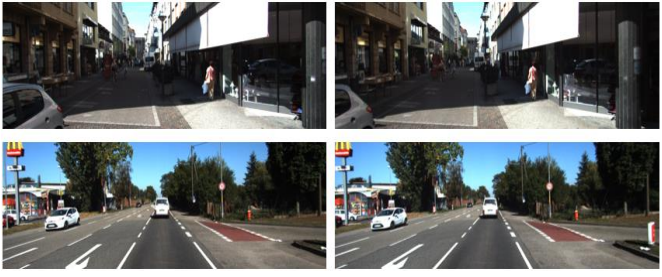

**Figure 3.** Illustration of samples of stereo images in the KITTI dataset. Left image is the left vision, and the right image is the right vision.

the camera transformation matrix, which means that a point in the world coordination would be transformed to another due to the camera ego-motion. Then we can use the relationship between the image coordination and the world coordination to transform the world coordination to the image coordination, and then construct the target image by all the source images. The mathematic description of the process is as follows:

Given the camera intrinsic matrix $K$, we can transform the image coordination of each pixel coordinate $(u_t, v_t)$ in image $I_l^t$ to the world coordinates $(X_t, Y_t, Z_t)$:

$$\begin{bmatrix} X_t \\ Y_t \\ Z_t \\ 1 \end{bmatrix} = Z_t K^{-1} \begin{bmatrix} u_t \\ v_t \\ 1 \end{bmatrix} \quad (4)$$

Then with the camera 6DoF motion estimation matrix $[R\ T]$, we can get its 3D position in the next moment:

$$\begin{bmatrix} X_{t+1} \\ Y_{t+1} \\ Z_{t+1} \end{bmatrix} = [R\ T] \begin{bmatrix} X_t \\ Y_t \\ Z_t \\ 1 \end{bmatrix} \quad (5)$$

In this way, we can get the reconstructed image $I_l^{t+1}$ by using the photometric measures $I_l^t(u_t, v_t) = I_l^{t+1}(u_{t+1}, v_{t+1})$, where

$$Z_{t+1} \begin{bmatrix} u_{t+1} \\ v_{t+1} \\ 1 \end{bmatrix} = K \begin{bmatrix} X_{t+1} \\ Y_{t+1} \\ Z_{t+1} \\ 1 \end{bmatrix} \quad (6)$$

Thus, the temporal reconstruction loss can be computed by:

$$\mathcal{L}_{vs} = \sum_s \sum_p |I_l^t(p) - \hat{I}^s(p)| \quad (7)$$

where $p$ index the pixel coordinates in $\{u_t, v_t\}$, $\hat{I}^s(p)$ is the reconstructed target image from the source image $I^s(p)$. Then depth can be estimated as an intermediate quantity.

Notice that the projected coordinates $\{u_{t+1}, v_{t+1}\}$ are not discrete values as real image pixel. To obtain $I^s(u_{t+1}, v_{t+1})$ for populating the value of the reconstructed image $\hat{I}_s(p)$, we use the differentiable bilinear sampling method proposed in the spatial transformer networks [20], linearly interpolates the values of the nearest neighbors of $\{u_{t+1}, v_{t+1}\}$ to approximate $I_s(u_{t+1}, v_{t+1})$. If both $u_{t+1}$ and $v_{t+1}$ coordinates need to be discretized, we choose the nearest four neighbors top-left, top-right, bottom-left, and bottom-right. For the cases that either $u_{t+1}$ or $v_{t+1}$ coordinate is integer, we choose 6 neighbors by adding other nearest two points. Thus we can easily get the reconstructed images by all source images.

In this paper, we combine these two ways together to rebuild each target frame in a video. We use the two reconstruction losses simultaneously to predict both the disparity and the pose. Thus the total reconstruction loss for our method can be expressed as:

$$\mathcal{L}_{syth} = \sum_t \mathcal{L}_{vs}^t + \lambda_a \mathcal{L}_{ap}^t + \lambda_c \mathcal{L}_{lr}^t \quad (8)$$

where $\lambda_a$ and $\lambda_c$ are the weights for the left-right image reconstruction loss and the disparity consistency regularity, respectively. Note that the idea of view synthesis as supervision for learning single-view video depth is popular in recent years. However, to the best of our knowledge, all previous works require to adjust the range of depth to the internal [0, 10] to make the depth and pose estimation more feasible, while our framework can be applied to directly predict the video depth without auxiliary assumption. It is worth mentioning that our method can achieve even better results by only using 1/2x training data.

*B. Depth Estimation as Image Reconstruction by both ego-motion and disparity consistency*

As we have shown previously, the image synthesis can be implemented with the fully differentiable loss functions with CNNs in the geometry and pose estimation modules. As indicated in Eq. 8, the key component of our framework is a differentiable disparity image-based target view image reconstruction by sampling pixels from a source view based on the predicted disparity map, camera motion estimation and the relative pose of the two cameras.

The above inter-frame view synthesis formulation is feasible for monocular videos which is mainly based on the assumptions that the scene is static without moving object, the vision difference is caused by the camera pose change and no new object appears into the view between the target view and the source views. But this is hard to satisfy for all training sequences collected in real world. To overcome this obstacle, an additional mask network is necessary to filter out these undesirable factors that lead to instability of training. This part is trained simultaneously with the depth and pose networks, and outputs per pixel mask $\hat{M}_s(p)$ for each target-source pair, indicating where the view synthesis can be successfully modeled for each target pixel. Thus the inter-frame reconstruction loss can be rewritten as:

$$\mathcal{L}_{vs} = \sum_r \mathcal{L}_{vs}^r = \sum_r \sum_s \sum_p \hat{M}_s^r(p) |I_l^t(p) - \hat{I}_s(p)|. \quad (9)$$

Where r indexes over different feature scales of the image, s indexes over the source images.

Training with the above loss would always tend to predicting $\hat{M}_s(p)$ to be zero, which perfectly minimizes the loss. To solve this problem, we use the regularization term $\mathcal{L}_{reg}(\hat{M}_s^r) = -\sum_{i,j} \ln(\hat{M}_{s,i,j}^r)$ as in [14] to encourage nonzero predictions. In other words, the network is encouraged to minimize the view synthesis objective, but allows certain slackness for discounting the factors not be considered by the model.

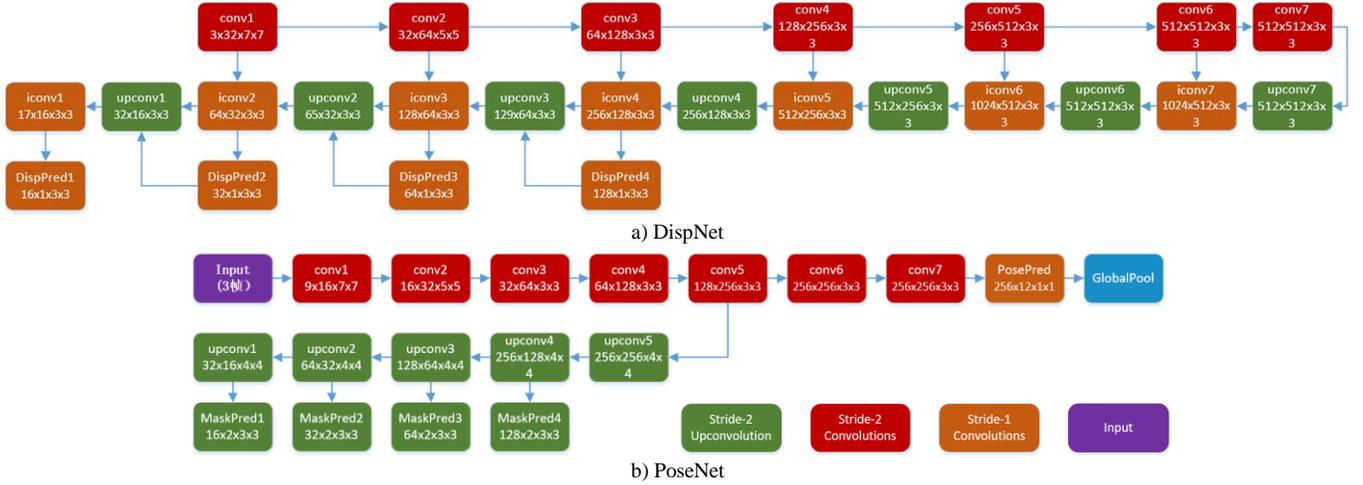

**Figure 4.** Network architecture for our depth, pose and mask prediction modules. In a) shows the disparity network structure and b) shows the pose network structure combined with the motion mask module.

One remaining issue is that the gradients are mainly derived from the pixel intensity difference between $I_t(u_t, v_t)$ and the four neighbors of $I_{t+1}(u_{t+1}, v_{t+1})$, which would be hard to train if the correct $\{u_{t+1}, v_{t+1}\}$ is located in a low-texture region or far from the real value. This is a well-known issue in motion estimation. We adopt explicit multi-scales prediction and smoothness loss as in [14, 18] that allows gradients to be derived from larger spatial regions directly in this work. We encourage disparities to be locally smooth with an L1 penalty on the disparity gradients. As depth discontinuities often occur at image edges, similar to [18], we use an edge-aware smoothness loss term by taking the image gradients into account.

$$\mathcal{L}_{smooth}^t = \Sigma_{u,v} |\partial_x \partial_x d_{u,v}^t| e^{-|\partial_x I_{u,v}^t|} + |\partial_y \partial_y d_{u,v}^t| e^{-|\partial_y I_{u,v}^t|} \quad (10)$$

Then our final loss function becomes:

$$\mathcal{L}_{final} = \mathcal{L}_{syth} + \Sigma_t \lambda_s \mathcal{L}_{smooth}^t + \lambda_e \Sigma_s \mathcal{L}_{reg}(\hat{M}_s(p)). \quad (11)$$

where $s$ indexes over source images, and $\lambda_s$ and $\lambda_e$ are the weighting for the depth smoothness loss and the mask regularization, respectively.

### C. *Network architecture of depth* and pose

**Depth network:** For unsupervised single-view depth prediction, we use a fully convolutional deep neural network which is similar to the supervised DispNetS architecture proposed in Mayer et al. [19], that is mainly based on an encoder-decoder design with skip connections and multi-scale predictions (see Fig. 4-a). For the prediction layers, we use 1/sigmoid(x) to make the predicted depth to be always positive within a reasonable range. The networks output their results at 4 different spatial scales. As the multi-scales features are up-sampled by the factor of 2, the disparity of neighboring pixels in each scale will differ due to the scaling factors. To correct this problem, we scale the disparity smoothness term with a factor r for each scale level to get equivalent smoothing at each level. Thus $\lambda_s = 0.1/r$, where r is the downscaling factor of the corresponding layer based on the scale of the input image to the network.

**Pose Network:** The input to the pose estimation network is the target view concatenated with all the source views along the channels, and the outputs are relative motions between the target view and each of the source views. The network consists of 7 convolutional layers followed by a $1 \times 1$ convolution with $6 * (N - 1)$ outputs channels, corresponding to rotation angles and translations along the coordinate axis. Finally, global average pooling is applied to obtain the predictions.

The mask prediction network shares the first five feature encoding layers with the pose network, followed by 5 de-convolutional layers. The number of output channels for each prediction layer are $2 * (N - 1)$, with every two channels normalized by softmax to obtain the mask prediction for the corresponding source-target pair and the second channel after normalization is used for computing the smooth loss in equation (10). The networks can output their results at 4 different spatial scales and the largest scale is used as the predicted scale.

### IV. EXPERIMENTS

In this section, we evaluate the performance of our approach and compare with prior approaches on unsupervised monocular depth as well as supervised monocular depth estimation mainly on the KITTI dataset [21]. We also use the Make3D dataset [1] for evaluating cross dataset generalization ability.

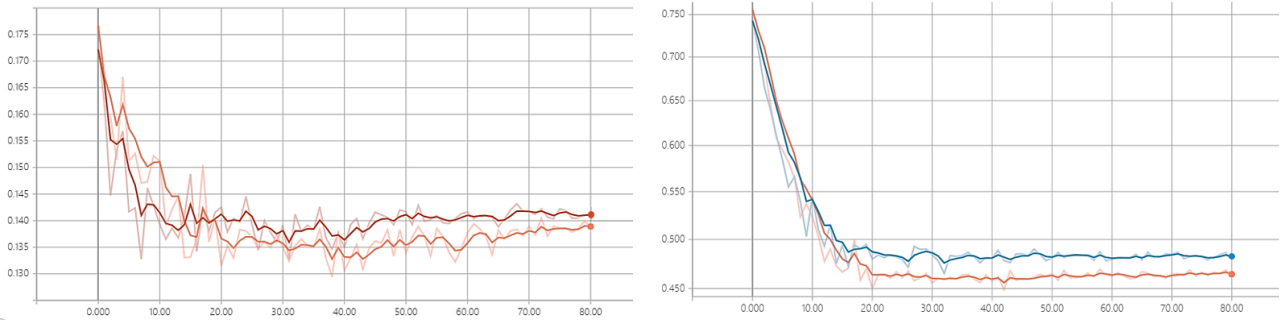

**Figure 5. Left:** Comparison of the changing curve of the frame-to-frame image reconstruction loss on validation set between Zhou *et al*. (red) [14] and ours (orange). **Right:** Comparison of the changing curve of the stereo image reconstruction loss on validation set between Godard *et al*. (blue) [10] and ours (orange).

**Table 1:** Results on KITTI dataset [21] using the split of Eigen [4] compared with supervised methods and the baseline with the results provided by authors, where lower is better for the error metric, and higher is better for accuracy metric.

| Method | Error metric | | | | Accuracy metric | | |
|---|---|---|---|---|---|---|---|
| | Abs Rel | Sq Rel | RMSE | RMSE log | $\delta < 1.25$ | $\delta < 1.25^2$ | $\delta < 1.25^3$ |
| Eigen et al.[4] | 0.203 | **1.548** | **6.307** | 0.282 | 0.702 | 0.890 | 0.958 |
| Liu et al.[2] | 0.202 | 1.614 | 6.523 | 0.275 | 0.678 | 0.895 | 0.965 |
| Zhou et al.[14] | 0.208 | 1.768 | 6.856 | 0.283 | 0.678 | 0.885 | 0.957 |
| Ours | **0.195** | 1.754 | 6.505 | **0.271** | **0.717** | **0.898** | **0.961** |

**Table 2:** Results on KITTI dataset [21] using the split of Eigen [4] compared with unsupervised method and the baseline by our predictions

| Method | Error metric | | | | Accuracy metric | | |
|---|---|---|---|---|---|---|---|
| | Abs Rel | Sq Rel | RMSE | RMSE log | $\delta < 1.25$ | $\delta < 1.25^2$ | $\delta < 1.25^3$ |
| Zhou et al.[14] | 0.207 | 2.333 | 6.658 | 0.282 | 0.670 | 0.883 | 0.954 |
| Godard et al.[10] | 0.206 | 1.910 | 7.943 | 0.320 | 0.696 | 0.852 | 0.923 |
| Ours | **0.195** | **1.754** | **6.505** | **0.271** | **0.717** | **0.898** | **0.961** |

### A. Implementation Details

The algorithm was implemented in the PyTorch [22] framework. For all the experiments, we set the weighting of the different loss components to $\lambda_s = 0.2, \lambda_e = 0.2, \lambda_a = 0.5$ and $\lambda_c = 0.5$. During training, the batch normalization [23] was used for all the layers except the output layers. We trained our model from scratch for 80 epochs, with the Adam [24] optimizer, Gaussian random initialization and mini-batch size of 4. We optimized all four scales at once led to more stable convergence. Similarly, we found that weighting them differently would lead to unstable convergence.

The network takes almost 32 hours to train using a single 1080Ti GPU on a dataset of 16 thousand images for 80 epochs, the changing curve of the training error is shown in Fig. 5. The runtime of our model is beyond real time at test and takes less than 35 ms.

The learning rate was initially set to 0.0001 and halving it every 10 epochs until the end. The network was first pre-trained on the larger Cityscapes dataset [25], and then fine-tuned on KITTI, which resulted in slight performance improvement. All the experiments were performed with image sequences captured by both color monocular cameras with fixed focal length. We resized the images to $512 \times 256$ during training, but the network can be tested with arbitrary input image size, due to both the depth and pose networks with the fully-convolutional structure.

We fixed the length of image sequences to be 3 frames, and selected the left video central frame as the target views, their counterpart right images and the ±1 frames as the source views. We only used the left stereo image as input and the right stereo image just used during training to reconstruct the target image. We trained our system on the split dataset provided by [4] which excluding all the frames from the testing scenes as well as static sequences for training. This results is trained in a total of 22,801 sequences where **16,384** for training and **5,730** for validation, compared with the baseline method which needs 40109 sequences.

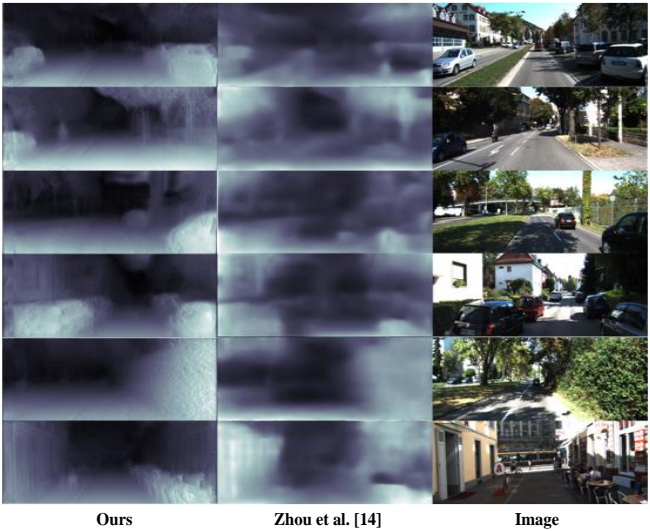

**Figure 6.** Comparison of single-view depth predictions on the KITTI dataset between Zhou *et al*. [14] (with only temporal image synthesis, and the smooth loss ignoring the image edges), and ours (with both spatial and temporal image synthesis and edge aware smooth loss).

Data augmentation is used during training. We flip the input images horizontally with a 50% chance, taking care to also swap both images so they are in the correct position relative to each other. We also added color augmentations, with a 50% chance, where we performed random gamma, brightness, and color shifts by sampling from uniform distributions in the ranges [0.8,1.2] for gamma, [0.5,2.0] for brightness, and [0.8,1.2] for each color channel separately as in [10].

*B. Single monocular image based depth estimation*

**KITTI:** In its raw form, the dataset contains 42,382 rectified stereo pairs from 61 scenes, with a typical image being 1242×375 pixels, only few images have subtle differences. We present the single-view depth results for the KITTI dataset [21] using the 697 images from the test split of [4] which covers a total of 29 scenes, to enable comparison with existing works.

To the best of our knowledge, no previous approaches exist that learn single-view depth from monocular videos in an unsupervised manner combined with stereo reconstruction loss. Nonetheless, here we provide comparison with prior methods both on baseline method [14] unsupervised depth estimation [10] and supervised depth estimation [2, 4]. First we compare the different smooth losses influenced for our method, and find that the second-order gradient smooth of the depth map with image first order slack term has the competitive performance with first-order depth map edge gradient smooth with the same image edge slack operation. For fair comparison, we use the same split manner as in [10] and evaluate the prediction with the same resolution as the input image. Fig.6 provides some examples of visual comparison between our results and the baseline method [14] over a variety of examples. We can see that although our model is trained by unsupervised manner, our results are comparable to that of the supervised methods, and can preserve the depth boundaries and thin structures such as trees and street lights better. As shown in Table 1, our unsupervised method performs comparably with several supervised methods (e.g. Eigen *et al*. [4] and Liu *et al*. [2]), the results of the compared methods are recomputed by us given the authors' original predictions to ensure that all the scores are directly comparable. In Table 2, we have shown the results compared with several unsupervised depth estimation method which is computed by ours. The results are slightly different from the author provided within accepting error due to the different training model. For all the compared methods we use bilinear interpolation to resize the prediction results to the input image size.

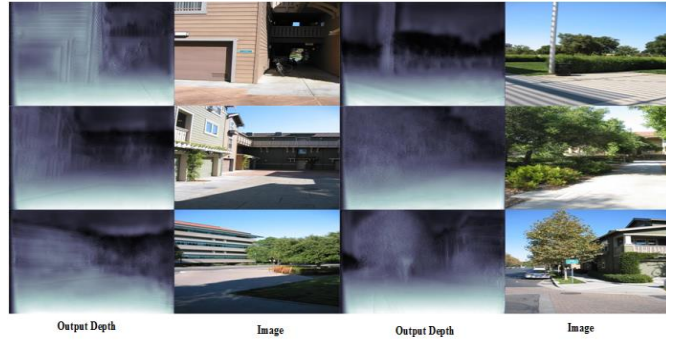

**Figure 7.** Illustration of the samples of depth predictions on the Make3D dataset. Note that our model is only trained on KITTI dataset, and directly tested on Make3D.

**Resnet 18 and Resnet 50**

For the sake of completeness, as similar to [10], we also show a variant of our model using Resnet50 [26] as the encoder, the rest of the architecture, parameters and training procedure staying identical. But we find that it does not produce a significant improvement, we guess this is caused by the lack of efficient training data.

**Post-processing**

We perform a post-processing step on the output to reduce the effect of stereo dis-occlusions which create disparity ramps on both the left side of the image. For an input image at test time, we also compute the disparity map for its horizontally flipped image. By flipping back this disparity map we obtain a disparity map where the disparity ramps are located on the right side of the image. We combine both disparity maps to form the final result by assigning the first 5% on the left of the image using the original disparity and the last 5% on the right to the flipped disparities. The central part of the final disparity map is the average of the two. This final post-processing step leads to an accuracy increase of 0.1% and less visual artifacts at the expense of doubling the amount of test time computation. We guess this is because of

**Generalizing to Make3D dataset**

To illustrate the generalization ability of our method, we performed our model on the Make3D test set of [1]. Make3D consists of only RGB/Depth pairs and no stereo images, thus our method cannot be trained by these data. We use our network trained only on the KITTI dataset, despite the dissimilarities of the datasets, both in contents and camera parameters, we still achieve feasible results. Qualitative results are shown in Fig.7, these results would likely be improved with more relevant training data.

### 4.6. Limitations

Even though our model effectively improves the quality of the results, there are still some artifacts visible at occlusion boundaries due to the pixels in the occlusion region not being visible in both images. Our method requires rectified and temporally aligned stereo pairs during training, which means that it is currently not possible to use the existing single-view datasets for training purposes. However, it is possible to fine-tune our model on application specific ground truth depth data. Finally, our method mainly relies on the image reconstruction term, meaning that specular region and transparent surfaces will produce inconsistent depths. This could be improved with more sophisticated similarity measures.

## V. CONCLUSION

In this paper, we have presented an end-to-end unsupervised depth learning pipeline that adds the additional constraints of left and right calibrated stereo image frames reconstruction to the existing view synthesis based unsupervised single-view video depth estimation. This system is trained by unlabeled videos and its counterpart stereo view image, but just needs one image as input. In this way, our method can reduce the error of depth estimation and perform comparably with approaches that trained by ground-truth depth or known pose. The edge relaxation L2 smooth regularization term is also used to make the prediction more nature and feasible, and the obtained depth estimation has better contour for the foreground target. Our method can achieve competitive results with the baseline method in the case of only by almost 3/5 times samples and need not to capture expensive ground truth depth. The experimental results also show that our method can generalize well to the unseen datasets.

Despite good performance on the benchmark evaluation, a number of challenges are yet to be further addressed: 1) our current framework does not explicitly estimate scene local motions and occlusions which is critical for 3D scene understanding. We can solve it by modeling of scene dynamics through motion segmentation; 2) our framework assumes the camera intrinsic are known, which forbids the use of arbitrary Internet videos collected by unknown cameras, we plan to address this in future work; 3) another interesting area for future work would be to investigate the portability of our algorithm to mobile embedded systems using the recent popular network compression technology.